# Design, modelling and control of a novel agricultural robot with interlock drive system

David Reiser[1], Volker Nannen[2], Gero Hubel[1], Hans W. Griepentrog[1]

*Abstract*— A current problem in the design of small and lightweight autonomous agricultural robots is how to create sufficient traction on soil to pull an agricultural implement or load. One promising solution is offered by the interlock drive system, which penetrates spikes into the soil to create traction. The combination of soil penetrating spikes and a push-pull design offers new possibilities for vehicle control. By controlling the interlocking of the spikes and pushing and pulling them against the main frame, the vehicle can perform tight maneuvers. To validate this idea, we designed a robot, capable of creating high traction and performing headland turns. The navigation of the new robot system is performed by actively pushing the spikes, mounted on a slide into the soil, while the main frame is pushed back and pulled forward. The vehicle of 2-meter length was able to turn on the spot, and could follow a straight line, just using the spikes and the push-pull mechanism. The trajectory and the performed measurements suggest, that a vehicle which uses only spikes for traction and steering is fully capable of performing autonomous tasks in agriculture fields.

## I. INTRODUCTION

The increase of world population and simultaneously rising demand on high quality food, will bring a lot of challenges for future farming. Robotics could bring high benefits for agriculture in performing autonomous tasks for soil cultivation, seeding, weeding, plant care and harvest. Today agricultural machines are getting bigger and bigger to save labor force for the driver [1]. The resulting heavy weight machines compacting the soil, which limits soil aeration, water infiltration and root penetration [2]. This would not be necessary any more, when autonomous robots could perform the tasks. Because of safety also small and light machines are preferable, malfunction and hazards would not cause a catastrophe, where humans would get hurt. Additionally, the robots could minimize the soil compaction [2], [3].

Small robots with conventional wheeled drive systems can just apply a small amount of their own weight into traction force, enabling the system just to create small interaction forces [4]. When preparing the soil e.g. before seeding, high interaction forces are needed, which would limit the minimum size of machines. An alternative traction without compacting the soil would be to interlock with the terrain, using spikes or hooks. This was first proposed by a patent of Bover et. al. [5]. The principal idea is to interlock spikes into the soil and pull the machine over the spikes.

[1] *Institute of Agricultural Engineering, University Hohenheim, Garbenstr. 9, D-70599 Germany; correspondence: (david.reiser) at uni-hohenheim.de*
[2] *Institute of Mobile Systems Engineering, Koblenz-Landau University, Koblenz 56070, Germany*

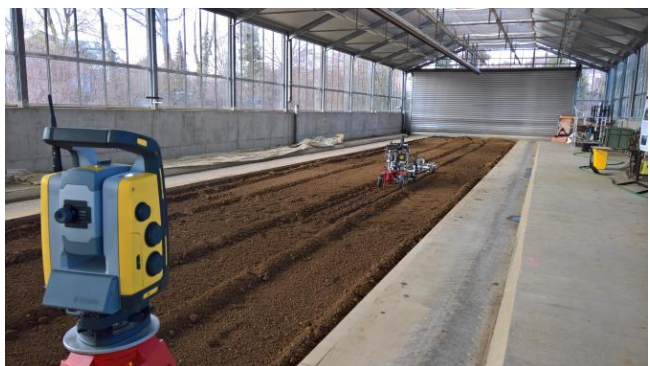

Fig. 1: The developed interlock drive system in action, tracked by a total station in the soil bin.

In research there are many different methods to increase the interaction force with the environment. Typical methods include climbing, nano- and micro-robots. Some use hooks to create more traction, others use adhesive pads to create high interaction forces [6], [7]. Many use the example of insects like ants to create controllable adhesives, enabling them to create interaction forces many times their body weight [8]. Others use ploughing for decreasing slip on steep slopes [9]. Different methods improved the kinematics or the traction of tires. Also spiked or blade tires were used for agricultural purposes to improve the force transmission to the soil [10]. On a field, the necessary interaction force is varying immense over the field. Therefore, a mechanic is necessary what could automatically adapt to the estimated draught force, just interlocking with the soil exactly at the necessary depth. At the same time the system should be light weight, to minimize the soil compaction and be able to create sufficient traction to pull an implement or load.

The present article evaluates a novel robot design which uses active and passive interlocking of spikes with the soil. It is able to follow straight rows, could steer and is able to turn at the end of the field (headland turn). The system is light weight, to minimize the soil compaction while being able to create sufficient traction to pull an implement or load. The target application for the prototype presented here is to perform mechanical weeding in crop rows.

The paper is organized as follows: First, the principle of the locomotion and the overall approach is described, how the spikes could interlock with the soil and how the robot used the method to steer. Second, the mechanic and control structure of the robot is described. Afterwards, the performed tests and the experiment environment are described, before the results are depicted and everything gets concluded. As last point future work is mentioned.

## II. ROBOT DESIGN AND DEVELOPMENT

### A. The overall approach and idea

The main idea is to design a set of passive articulated spikes which are connected to a lever and a joint at the robot. Pushed in one direction, the spikes penetrate into the soil under their own weight, the force transfer from the joint, and the resistive force of the soil. Pulled in the other direction, they are extracted out of the soil by a reversal of the forces in the soil and at the joint. This is a self-regulating passive design: depending on the draft force required by for example a tillage implement, the spikes automatically adjust their depth in the soil to generate the required pull.

The spikes are combined with a push-pull vehicle design where two frames are moved with respect to each other, such that one part of the vehicle moves forward while the other part provides traction with the spikes. Our push-pull vehicle consists of a main frame and a slide which can move along the central axis of the frame but cannot rotate or shift laterally with respect to the central frame. At any one time, the slide acts as the tractive device which interlocks with the soil by penetrating one or more spikes into the soil, while the main frame acts as the implement carrier which moves forward while pulling for example an implement through the soil.

Our design target for the robot was to navigate through crop rows in a maize field and to perform agricultural tasks like mechanical weeding. Maize rows are typically planted 0.75 m apart. A margin of 0.1 m per side is needed for steering, which limits the width of the robot to 0.55 m. The length of the robot was limited to 2-meter length, for ease of transportation.

Since the assumption was that when just one of the spikes would enter the soil, the momentum between the center of mass and the spike would steer the robot, the spikes where mounted at the outer ends of the movable slide. To allow full control over the motion of the robot by using only the forces on the spikes and any application tools, the main frame was mounted on castor wheels.

To allow a passive and active mode of the two spikes, a linear actuator was attached to the freely movable spike, connected to a lever. When the actuator was completely contracted, the spike was pulled out of the soil. In half extended position the spike was free to penetrate under the power of a backward motion. Further extension of the linear actuator initiated an active penetration of the spike into the soil to interlock the spike and guarantee traction. The three modes of the linear actuator could be seen in the following Fig. 2. The spikes could be controlled individually based on the needs of the trajectory of the robot.

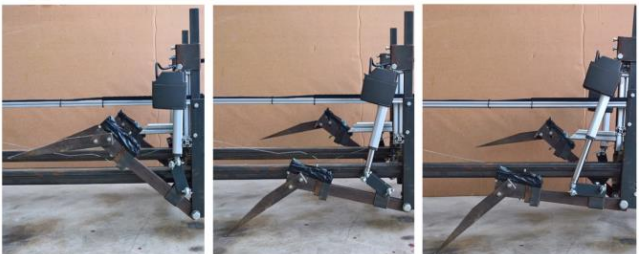

Fig. 2: Active and passive mode of the spikes mounted to the slide.

### B. Steering principle

Our working hypothesis was that the internal kinetics of the motion between a frame and slide as described above can be used to steer the vehicle. The fundamental principle here is that both parts have a center of motion resistance with respect to the ground, which can be calculated from the location and resistance of all contact points of each part with the ground. Any force applied at the motor is translated into a force that either pushes those two centers apart or pulls them together.

If the two centers are on a line parallel to the central axis of the main frame, any force that pushes or pulls those two centers together or apart will simply move them along the central axis and the vehicle will move forward in a straight line. This follows directly from the fact that the slider can move along the central axis of the main frame, but cannot rotate or move laterally with respect to the main frame,

What happens when the centers are not aligned with the main axis? For simplicity we will only consider the case where only the left or the right spike of the slide penetrates into the soil, such that the center of resistance of the slide is exactly at the spike. Assuming that the spike is firmly anchored such that it can rotate in the soil but not move laterally, the center of resistance of the main frame is now pulled towards or pushed away from this single spike. As shown in Fig. 3, this creates a rotational moment at the slide about the spike, forcing the vehicle to turn.

The resulting motion of the main frame during such a turning maneuver is best understood by considering the center of resistance of the main frame $A$ and a point $C$ where a line through $A$ and parallel to the central vehicle axis intersects with a perpendicular line through the anchored spike. Let $s_1$ and $s_2$ be the distance from this point to the anchored spike, and let $x$ be the changing distance from $C$ to the center of resistance of the main frame, such that during contraction $x_1$ is the value of $x$ when the vehicle is fully extended, and $x_2$ is the value of $x$ when the vehicle is fully contracted. These lengths define two angles $\gamma_1 = \mathrm{atan}(s_1/x_1)$ and $\gamma_2 = \mathrm{atan}(s_1/x_2)$. Assuming that the left spike is anchored, the angle $\alpha$ by which the main frame rotates during a contraction from $x_1$ to $x_2$ is

$$\alpha = \mathrm{atan}\left(\frac{s_1}{x_1}\right) - \mathrm{atan}\left(\frac{s_2}{x_2}\right). \tag{1}$$

This is a clockwise rotation and the angle α will be negative. If the right spike is anchored instead, the vehicle will turn counter-clockwise and the angle will be positive.

The calculation of rotation during contraction is in principle equivalent, but at this point we need to take the dynamic weight transfer of the vehicle into account. Dynamic weight transfer occurs because the main frame moves on rollers which interact with the soil at the surface, while the spikes penetrate the soil and interact with the soil below the surface. When the push-pull mechanism contracts, a spike effectively pulls at the rear rollers from below, increasing their friction with the ground, and pushes the front rollers from below, decreasing their friction with the ground. This moves the geometric center of resistance of the main frame backwards towards the rear rollers. Weight transfer also moves the center of resistance laterally, changing the value of $s$ by an amount $z$. During expansion the forces are inverted, and the geometric center is pushed a few centimeters forwards towards the front rollers. Let $y_1$ and $y_2$ be the distance along the central vehicle axis between the center of resistance of the main frame at contraction and expansion at the extreme points of the motion cycle and let $z_1$ and $z_2$ be the corresponding distances parallel to $s$. Let $x_3 = x_1 - y_1$ be the value of $x$ during expansion when the vehicle is fully expanded, and $x_4 = x_2 - y_2$ be the value of $x$ when the vehicle is fully contracted. Let $s_3 = s_1 - z_1$ and $s_4 = s_2 - z_2$. We can now calculate the angle $\beta$ during contraction as

$$\beta = \operatorname{atan}\left(\frac{s_3}{x_3}\right) - \operatorname{atan}\left(\frac{s_4}{x_4}\right). \quad (2)$$

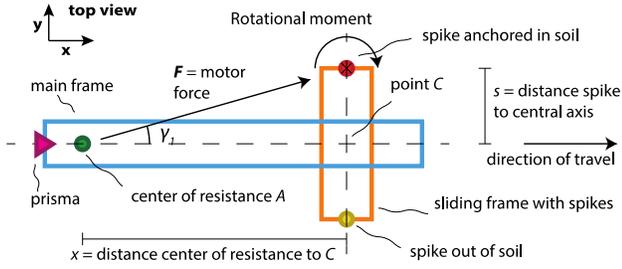

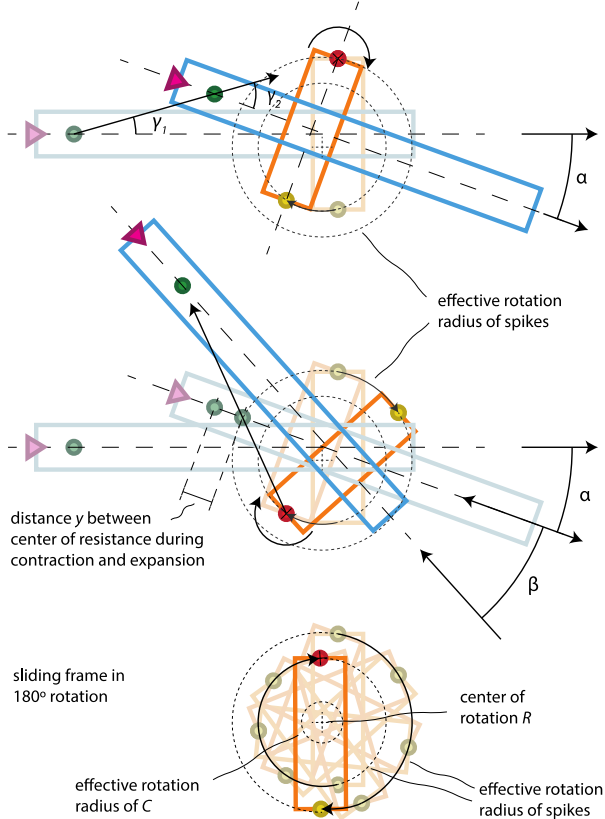

Fig. 3: Graph explaining the forces for turning the vehicle between frame and slide.

The beauty of Equations (1) and (2) is that the turning angle can be calculated if the position of $A$ is known at the beginning and end of each contraction, avoiding the need for integration over curves. Regarding the location of $A$ we find that $x$ and $s$ are design parameters and that their deviation in the field which is completely described by the distances $y_1$, $y_2$, $z_1$ and $z_2$. Those deviations depend on $x$ as well as the penetration depth of the spikes and the soil resistance at the castor wheels and should be calibrated in the field. However, given the weight and the design dimensions of the present vehicle, we can calculate first approximations: the angle $\alpha$ should be 21 degrees and the angle $\beta$ should be 32 degrees, such that one push-pull cycle with only a left or right spike in the ground can turn the vehicle by about 53 degrees, which will be tested in the experimental section.

Concerning the overall center of rotation $R$ of the vehicle during the turning maneuvers, during each contraction and expansion, the center of rotation of the slide is either its right or left spike. For $\lim_{\alpha \to 0}$ and $\lim_{\beta \to 0}$ we find that the effective center of rotation will be in the middle between the two spikes. For non-trivial values of $\alpha$ and $\beta$ the center of rotation is a few centimeters to the right of this center when turning clockwise, and a few cm to the left of $C$ when turning counterclockwise, and this distance is larger for larger values of $\alpha$ and $\beta$. The exact radius of $C$ around $R$ also depends on the slip of the spikes during operations (see Fig. 3). This will also be tested in the experimental section.

## III. SYSTEM INTEGRATION

### A. Mechanical design

The mechanical design of the robot was based on a linear iron frame, able to move the slide back and forth along the central axis of the main frame. The actuation of the slide was based on a basic DC electrical motor with a nominal power of 120 W (BCI-63.55, emb-papst GmbH, Mulfingen, Germany) was connected with a chain to this slide. The spikes and the tool at the back were connected to linear motors LA25 of the company Linak (Linak, Nordborg, Denmark). They were able to create a push force of 1200 N, enough to lift up the whole vehicle. The overall weight of the vehicle was 90 kg. To enable the device to freely follow forces in all directions, the vehicle was placed on castor wheels. For position tracking a prism was attached at the highest part of the robot to track the actual position with a total station. Fig. 4 shows the position and the mechanical parts of the complete robot used for the experiments.

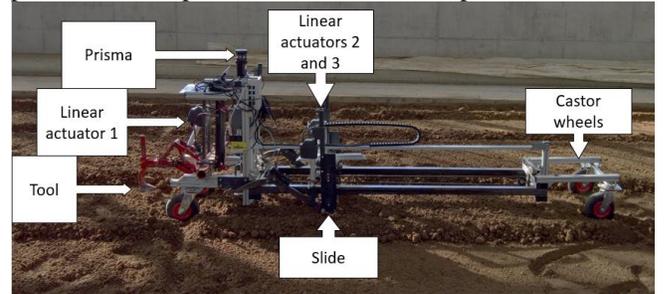

Fig. 4: Complete machine with linear actuators, tool and tracking prism and slide.

### B. Electronics, power consumption and autonomy

The whole system is controlled by a conventional tablet computer (HP, Atom x5 1.44 GHz, 4GB RAM), using

Ubuntu 16.04 and ROS Indigo [11]. The position of the robot was tracked by a total station using a Yuma Tablet and the control software of the supplier company Trimble. The used total station was a SPS930 (Trimble, Sunnyvale, CA, USA) with a tracking accuracy of +/- 4 mm in dynamical tracking mode. The orientation of the robot frame was tracked with a VN-100 Inertial Measurement Unit (IMU) (VectorNav, Dallas, USA) mounted to the main frame. The driving motor of the robot was controlled with a SDC2130 motor controller (Roboteq, Scottsdale, USA), including an encoder to log the actual movement of the slide. The linear motors were controlled using CAN bus protocol and with the use of a USB-CAN adapter.

The data of the total station was logged using a RS232 to USB adapter connected to the tablet computer. See Fig. 5 for an overview of the connections between the control tablet, the motors and the sensors.

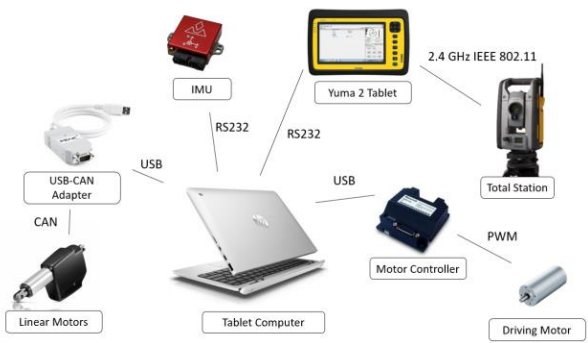

Fig. 5: Control, monitoring and data collection structure of the robot.

The motors and electronics were powered by two lead-acid batteries, providing a nominal voltage of 24 V and 17 Ah. The robot was controlled with ROS nodes using C++ and python languages. The data was logged in ROS-bag files and was later analyzed using Matlab R2017b. The position of the robot was gained by fusing the prism position and the IMU using the orientation of the robot. The tracking frequency of the prism was 5 Hz and the frequency of the IMU was collected with 50 Hz. For the actual position the data was interpolated. As the robot moved with a speed of 1 cm/s, the data rate for the position tracking was sufficient.

### C. Control strategy

For controlling the robot, three different methods were programmed. One method for running the motor straight, second method to move left and third method to move right. For moving straight, first the tool was lowered, the robot moved the slide to the front and pushed both spikes into the soil. Afterwards the slide was moved to the back of the robot, creating a forward movement. As soon as the vehicle was reaching the end of the frame, the spikes were pushed out of the soil and the frame was moved once more to the front. The turning of the motor direction was controlled using the encoder mounted at the chain of the slide. The chain travelled a distance of 1.12 m per contraction and expansion, but due to slip of the spikes when penetrating the soil, the vehicle travelled only about 1 m per cycle.

To turn the vehicle, we operated the spikes in active mode. One spike was pulled out completely, the actuator of the other spike was pushed down, and the vehicle pulled the spike towards the tool. At the end of the slide, the spike pushing into the ground was pulled up and the opposite spike was pushed down, followed by a slide movement to the front. The logged spike causes the robot frame to move backwards. To turn in the other direction, just the sides of the spikes were toggled. Each method could be individually defined in length and how often they should be repeated.

### D. Experimental setups

For testing the navigation of the robot, two different patterns were followed. One pattern was just moving straight for 5 cycles. The second pattern, was to turn left for 180 degrees. The calibration of the robot for the number of turning cycles needed were performed on concrete floor. Afterwards the same control program was tested on compacted soil in the University soil bin with a length of 46 m, 5 m width and 1.2 m depth. The total station was placed with clear line of sight for the whole test run and the complete setup from motor movement, motor torque and orientation was logged (see Fig. 1).

## IV. EXPERIMENT RESULTS AND DISCUSSION

The first tests on a concrete floor showed that the principle for navigation and steering worked, even on solid ground where the spikes cannot interlock. This is possible, because the linear actuators pushed the vehicle a little bit up to create friction force, high enough for steering. Empirical evaluation showed that a turn of 180 degrees needed 3 push-pull cycles, resulting in 60 degrees turn per cycle, well in line with our not calibrated calculation of 53 degrees in Section II.

The first test with 5 push-pull cycles of straight forward movement resulted in a total advance of 4.98 m. The coordinate position of the prism based on the total station, and the movements of the linear motors could be seen in the following Fig. 6.

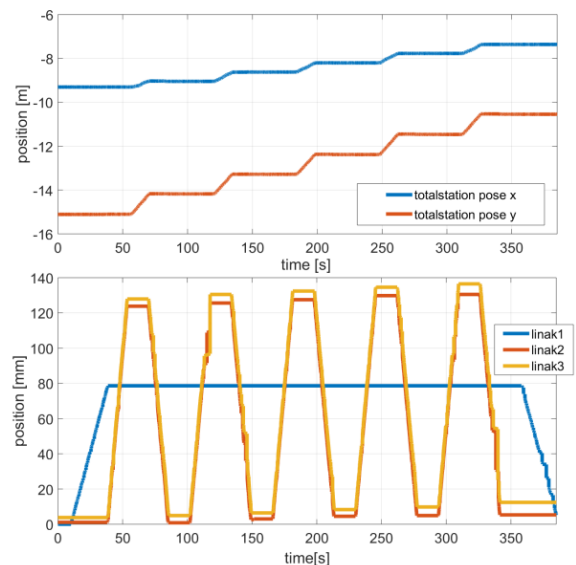

Fig. 6: Position of the robot and the linear motors following a straight line for 5 cycles.

First, the implement was lowered (linak1), afterwards both spikes were pushed down (linak2 and linak3), following a movement of the motor slide. Each cycle needed around 70 seconds, including the movement of the linear motors. The movement of the system followed small steps, as the robot was just moving forward for a short time period in the cycle. The counter force of the spikes was high enough to pull the tool at the implement even under compacted soil. It could be seen, that inhomogeneous soil compaction or higher forces caused the spikes to move deeper into the soil. The orientation of the robot system around the yaw axis was constant, as long as both spikes had enough grip, the robot followed a straight line, with small disturbances (see Fig. 7). Disturbances occurred based on soil disturbances and friction loss of the spikes, but were quite seldom.

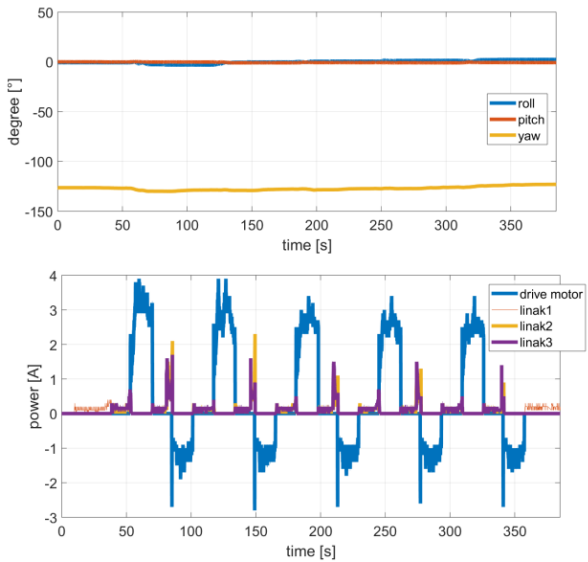

Fig. 7: Power consumption and orientations measured while following a straight line.

Therefore, it was assumed that the constant straight movement inside a crop row would be possible.

The electric consumption of the drive motor was dependent on the load. When pulling the implement forward, a mean current of 3 amperes was necessary for moving the vehicle. For moving the slide back to the front, a mean current of 1.5 A was sufficient. The linear motors consumed little power drawing intermittent peak currents which generally did not exceed one ampere. In straight movement the energy consumption of the vehicle was 75 W for the given weeding task.

When turning 180 degrees, the path of the robot was less clear than when following a straight line. The linear motor of the implement was moved up and the turning force was created based on the interlocked spikes in the soil. The position of the robot changed in x and y direction based on the actual driving direction and was accompanied by a constant turn. The position changed just slightly between start and stop of the turning mode (see Fig. 8). The total space needed for the turning based on the path tracked by the total station was 1.2 m in x and 1.9 m in y direction. Adding the shift of the slide to the maximum space of 1.12 m, a turning space of 3.02 m was needed for the vehicle. Therefore, the space needed for the whole vehicle to turn was quite small, compared to the size of the robot length. In the path an additional side shift could be seen for each direction change, not fitting with the theory. This was caused by the small lifting of the vehicle when the spike was hooked and turning behaviors of the castor wheels.

When looking at the turning rate for each cycle of the robot based on the IMU, it could be seen that the turning rate was not constant. The first cycle just caused the smallest turn, as the forces first had to turn the castor wheels of the robot. In general, the robot turned more degree, when the slide was moving to the front. A reason for this behavior could be the smaller friction and weight on the front of the robot, causing the vehicle to react easier to applied forces (see Fig. 9).

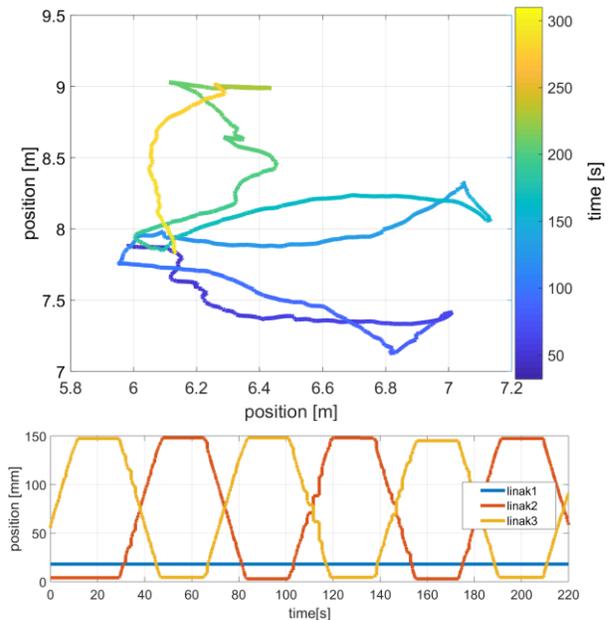

Fig. 8: Figure with results of the straight line, showing positions of the robot based on the time and the linear actuators.

Additionally, the shape of the spikes caused the robot to lift up the front when the vehicle was pushing the spike. When the spike was pulled to the center of mass, the spike shape caused the robot to press the front wheels into the soil, causing higher friction for this turn direction. The linear motors caused the robot to turn slightly around the roll axis, when the linear motors were pushed into the soil. This caused also better turning behavior as the center of mass tried to equal the roll behavior.

The forces of the drive motor were more constant than in the single forward movement, always around 2 A. This slightly higher load compared to unloaded movement (see Fig. 7) was caused by the torsion forces which were applied to the robot when just one of the spikes was hooked in the soil. The power consumption of the linear motors was also higher, as more force was needed when the spikes were pushed or pulled out of the soil. However, the overall power consumption was also here quite small, never exceeding more than 100 W consumption.

The tests showed that an interlock drive vehicle could be controlled, using a semi-active spikes with linear motors. The overall performance exceeded the expectations of the authors and indicates that agricultural work can perhaps be done with the high interaction forces generated by an interlocking drive mechanism. The nominal force applicable with a robot with interlock drive is much higher than the actual weight of the vehicle, causing low compaction to the soil. The power consumption was quite low, what would enable the system to work for long time independent on any power source.

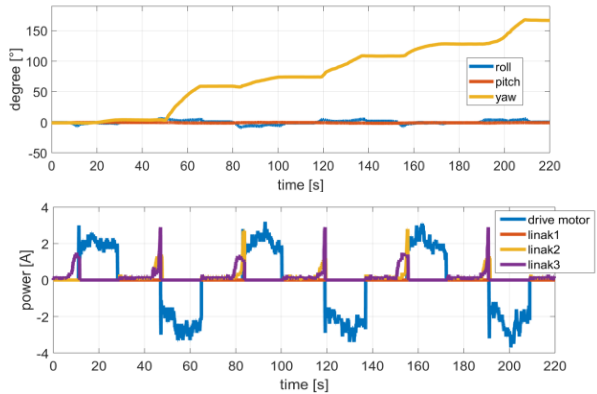

Fig. 9: Power and orientation changes for a turn of 180 degrees.

## V. Conclusion and Future Work

Our agricultural robot was able to pull an implement and to navigate using the novel principle of penetrating articulated spikes into the soil. This allowed the vehicle to move freely in orientation and position, controlled by a navigation algorithm and feedback system. A turn on the spot was possible and could be performed with 3 push-pull cycles and within a total time of 220 seconds. The degrees turned per push-pull cycle were in general agreement with the not calibrated calculations. We also found that the turning angle was much larger during expansion than during contraction, again in line with the theory.

Future work has to address the precise navigation of a vehicle under different soil and terrestrial conditions different slopes. Several shapes of spikes could be investigated, to address the best characteristics based on underground and environment variability. In combination with a solar panel the system could work in energy autarky. Combined with a battery pack the system could work easily 24 hours a day. Working in swarms, the machines could enable workloads of today's tractors or other agricultural machinery. As the system is driving slow with around 1 cm/s, it would stay at one spot anyhow for some time. This time could be used to analyze the soil with attached sensors, apply fertilizer or put seeds into the ground. Even using the interlocking spikes for taking soil samples could be possible. Therefore, the robot could be enabled to completely automate plant care in agricultural fields or enable new exploratory ways for robots. Fig. 10 is showing a vision graph depicting future work of the machine in an agricultural context.

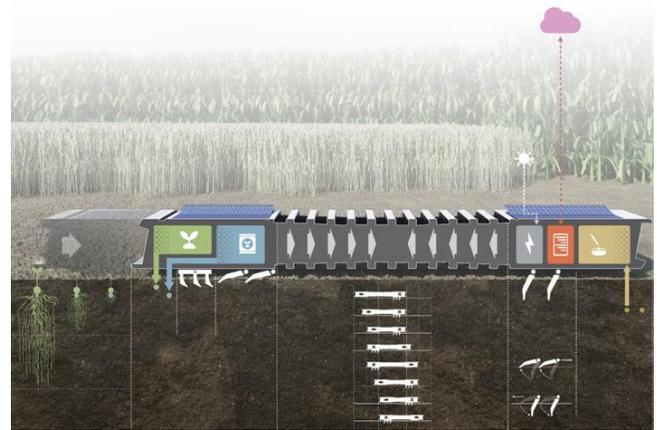

Fig. 10: Vision of the robot system based on the interlock drive system.

Acknowledgment

The authors wish to thank for the support of the companies Mädler and Linak for this work with materials and actuators.